\documentclass{article}
\usepackage{spconf,amsmath,graphicx}

\usepackage{lipsum}
\usepackage[dvipdfmx]{color}
\usepackage{bm,amssymb,mathrsfs,amsfonts}
\usepackage{caption}
\usepackage{scalefnt}
\usepackage{multirow}
\usepackage{enumitem}
\usepackage{type1cm}
\usepackage{cite}
\usepackage{algorithm}
\usepackage{algorithmic}
\usepackage{here}
\usepackage{lipsum}
\usepackage{tabularx}
\usepackage{url}
\usepackage{color}
\usepackage{soul}
\sethlcolor{yellow}

\newcommand{\argmax}{\mathop{\rm arg~max}\limits}

\title{Transfer learning of language-independent end-to-end ASR\\with language model fusion}
\name{Hirofumi Inaguma$^1{^*}$\thanks{*Part of the work reported here was conducted while the author was visiting Johns Hopkins University.}, Jaejin Cho$^2$, Murali Karthick Baskar$^3$, Tatsuya Kawahara$^1$, Shinji Watanabe$^2$}
\address{$^1$Graduate School of Informatics, Kyoto University, Kyoto, Japan\\
$^2$Johns Hopkins University, Baltimore, MD, USA
, $^3$Brno University of Technology, Czech Republic}
\begin{document}
\ninept
\sloppy 
\maketitle
\begin{abstract}
This work explores better adaptation methods to low-resource languages using an external language model (LM) under the framework of transfer learning.
We first build a language-independent ASR system in a unified sequence-to-sequence (S2S) architecture with a shared vocabulary among all languages.
During adaptation, we perform \textit{LM fusion transfer}, where an external LM is integrated into the decoder network of the attention-based S2S model in the whole adaptation stage, to effectively incorporate linguistic context of the target language.
We also investigate various seed models for transfer learning.
Experimental evaluations using the IARPA BABEL data set show that LM fusion transfer improves performances on all target five languages compared with simple transfer learning when the external text data is available.
Our final system drastically reduces the performance gap from the hybrid systems.
\end{abstract}
%


\section{Introduction}\label{sec:intro}
Fast system development for low-resourced new languages is one of the challenges in automatic speech recognition (ASR).
Recently, end-to-end ASR systems based on the sequence-to-sequence (S2S) architecture \cite{attention_nmt_bahdanau,attention_nips2015} are filling up the gap of performance from the conventional HMM-based hybrid systems and showing promising results in many tasks with its extremely simplified training and decoding schemes \cite{google_sota_asr,hybrid_ctc_attention,rwth_end2end}.
This is very attractive when building systems in new languages quickly.
However, models tend to suffer from the data sparseness problems in the low-resource scenario, especially in S2S models due to its data-driven optimization.

One possible approach to this problem is to utilize data of other languages.
There are various approaches to leverage other languages: (a) to train a model multilingually (multi-task learning with other languages), and then further fine-tune to a particular language \cite{dalmia2018sequence}, and (b) to adapt a multilingual model to a new language using transfer learning \cite{jj_slt18,tong2017multilingual,tong2018cross,dalmia2018sequence} and additional features obtained from the multilingual model such as multilingual bottleneck features (BNF) \cite{martin_analysis,language_independent_bnf,grezl2014adapting,vu2014multilingual} and language feature vectors (LFV) \cite{lfv} (cross-lingual adaptation).
To obtain a multilingual S2S model, a part of parameters can be shared while preparing the output layers per language \cite{dalmia2018sequence}, and we can further use a unified architecture with a shared vocabulary among multiple languages \cite{watanabe2017language,toshniwal2018multilingual,kim2018towards}.
Since it would take much time to train such systems from scratch for many languages  including new languages, we focus on the cross-lingual adaptation approach (b).

While a majority of the conventional transfer learning is concerned with acoustic model, using linguistic context during adaptation has not been investigated yet.
The research question in this paper is: \textit{Is linguistic context also helpful for adaptation to new languages?}
The most common approach to integrate the external language model (LM) is referred to as \textit{shallow fusion}, where LM scores are interpolated with scores from the S2S model \cite{kannan2017analysis,rwth_end2end,toshniwal2018comparison}.
Recently, several methods to leverage an external LM during training of S2S models are proposed: \textit{deep fusion} \cite{deep_fusion} and \textit{cold fusion} \cite{cold_fusion}.
In deep fusion, the decoder network in the pre-trained S2S model and an external Recurrent neural network language model (RNNLM) are integrated into a single architecture by the gating mechanism and only the gating part is re-trained.
In contrast, cold fusion integrates an external LM during the entire training stage.

In this paper, we investigate methods to fully utilize text data for adaptation to unseen low-resource languages.
We propose \textit{LM fusion transfer}, where an external LM is integrated into the decoder network of the S2S model only in the adaptation stage\footnote{Although we can perform LM fusion during training of the seed multilingual model, we focus on applying it during adaptation because our goal is to adapt it to a particular language rapidly.}, as an extension of cold fusion.
Since the decoder network is already well-trained in a language-independent manner, the model can better incorporate linguistic context from the external LM.
The extra cost to integrate the external LM during adaptation is trivial in the resource constrained condition.
We also investigate various seed multilingual models trained with 600 to 2200-hours speech data and show the effect of the amount and variety of multilingual training data.

Experimental evaluations on the IARPA BABEL corpus show that the LM fusion transfer improves performance compared to simple transfer learning with shallow fusion when the additional text data is available.
The performance of the transferred models is drastically improved by increasing the model capacity and incorporating the external LM, and the resulting models perform comparably with the latest BLSTM-HMM hybrid systems \cite{martin_analysis}.
To our best knowledge, this is the first results for the S2S model to show the competitive performance to the conventional hybrid systems in the low-resource scenario ($\sim$50 hours).



\section{Related work}\label{sec:related_work}
The traditional usage of unpaired text data in the S2S framework is categorized to four approaches: LM integration, pre-training, multi-task learning (MTL), and data augmentation.
In the LM integration approach, there are three methods: \textit{shallow fusion}, \textit{deep fusion}, and \textit{cold fusion} as described in Section \ref{sec:intro}.
Their differences are in the timing to integrate an external LM and the existence of additional parameters of the gating mechanism.
We depict these fusion methods in Fig. \ref{fig:fusion_transfer}.
In, \cite{toshniwal2018comparison}, these fusion methods are compared in middle-size English conversational speech ($\sim$300h) and large-scale Google voice search data.
However, none of previous works investigated the effect of them in other languages especially for low-resource languages, which is the focus of this paper.
In \cite{cold_fusion}, the authors show the effectiveness of cold fusion in a cross-domain scenario.
Since the external LM is more likely to be changed frequently than the acoustic model, it is time-consuming to train a new S2S model with the LM integration from scratch every time the external LM is updated.
In this work, we conduct LM fusion during adaptation to target languages, which is regarded as a more realistic scenario.

Another usage of the external LM is to initialize the lower layer in the decoder network with the pre-trained LM \cite{toshniwal2018comparison,lm_pretraining}.
However, we transfer almost all parameters in a multilingual S2S model (both encoder and decoder networks), and thus we do not explore this direction.
Apart from the external LM, the MTL approach with LM objective are investigated in \cite{toshniwal2018comparison,nmt_lm_mtl}.
Although the MTL approach does not require any additional parameters, it gets minor gains compared to LM fusion methods \cite{toshniwal2018comparison}.

Recently, data augmentation of speech data based on text-to-speech (TTS) synthesis is investigated in the S2S framework \cite{hayashi2018back,mumura2018slt}.
Since we are interested in the usage of linguistic context during adaptation, we leave this direction to the future work.


\section{End-to-end ASR}\label{sec:e2e_asr}

\subsection{Attention-based sequence-to-sequence}
We build all models with attention-based sequence-to-sequence (S2S) models, which can learn soft alignments between input and output sequences of variable lengths \cite{attention_nmt_bahdanau}.
They are composed of encoder and decoder networks.
The encoder network transforms input features $\bm{x}=(\bm{x}_{1},\dots,\bm{x}_{T})$ to a high-level continuous representation $\bm{h}=(\bm{h}_{1},\dots,\bm{h}_{T'})$ ($T' \leq T$), interleaved with subsampling layers to reduce the computational complexity \cite{las}.
The decoder network generates a probability distribution $P_{\rm S2S}$ of the corresponding $U$-length transcription $\bm{y}=(y_{1},\dots,y_{U})$ conditioned over all previous generated tokens:
\begin{align*}
& \bm{s}^{\rm S2S}_{u} = {\rm Decoder}(\bm{s}^{\rm S2S}_{u-1}, y_{u-1}, \bm{c}_{u}) \\
& P_{\rm S2S}(\bm{y}|\bm{x}) = {\rm softmax}(\bm{W}^{\rm o}\bm{s}^{\rm S2S}_{u} + \bm{b}^{\rm o})
\end{align*}
where $\bm{W}^{\rm o}$ and $\bm{b}^{\rm o}$ are trainable parameters, $\bm{s}^{\rm S2S}_{u}$ is a decoder state at the $u$-th timestep, and $\bm{c}_{u}$ is a context vector summarizing notable parts from the encoder states $\bm{h}$.
We adopt the location-based scoring function \cite{attention_nips2015}.
To encourage monotonic alignments, the auxiliary Connectionist Temporal Classification (CTC) \cite{ctc} objective is linearly interpolated \cite{joint_ctc_attention}.

During the inference stage, scores from the softmax layer used for the CTC objective are linearly interpolated in log-scale with a tunable parameter $\lambda \ (0 \leq \lambda \leq 1)$ to avoid generating incomplete and repeated hypotheses as follows \cite{hybrid_ctc_attention}:
\begin{align*}
\ln{P_{\rm ASR}(\bm{y}|\bm{x})} = (1 - \lambda) \ln{P_{\rm S2S}(\bm{y}|\bm{x})} + \lambda \ln{P_{\rm CTC}(\bm{y}|\bm{x})}
\end{align*}
\vspace{-6mm}

\begin{figure}[!t]
    \centering
    \includegraphics[width=0.90\hsize]{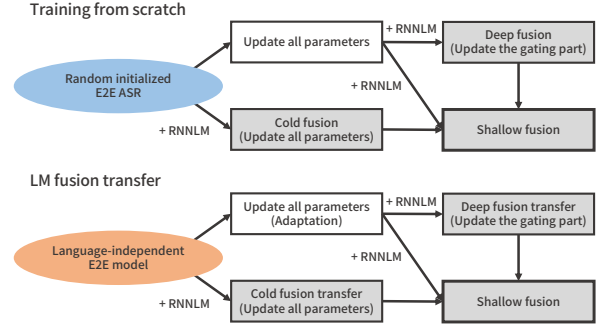}
    \vspace{-1mm}
    \caption{Overview of language model fusion transfer. LM fusion transfer is conducted with monolingual data only.}
    \label{fig:fusion_transfer}
    \vspace{-3mm}
\end{figure}

\subsection{LM fusion}\label{ssec:lm_fusion}
\subsubsection{Shallow fusion}\label{sssec:shallow_fusion}
In the conventional decoding paradigm with an external LM, referred to as \textit{shallow fusion}, scores from both the S2S model and LM are linearly interpolated to maximize the following criterion:
\begin{align*}
\bm{y^{*}} = \argmax_{\bm{y} \in \Omega^{*}} \{ \ln{P_{\rm ASR}(\bm{y}|\bm{x})} + \beta \ln{P_{\rm LM}(\bm{y})} \}
\end{align*}
where $\beta$ is a tunable parameter to define the importance of the external LM.
The separate LM, especially trained with a larger external text, has complementary effects to the implicit LM modeled in the decoder network.
Therefore, shallow fusion shows performance gains in many ASR tasks \cite{kannan2017analysis,rwth_end2end,toshniwal2018comparison}.

\subsubsection{Cold fusion (flat-start fusion)}\label{sssec:cold_fusion}
While shallow fusion uses the external LM only in the inference stage, \textit{cold fusion} \cite{cold_fusion} uses the pre-trained LM during training of the S2S model to provide effective linguistic context.
The fine-grained element-wise gating function is equipped to flexibly rely on the LM depending on the uncertainty of prediction:
\begin{align*}
& \bm{s}^{\rm LM}_{u} = \bm{W}^{\rm LM} \bm{d}^{\rm LM}_{u} + \bm{b}^{\rm LM} \\
& \bm{g}_{u} = \sigma(\bm{W}^{\rm g}[\bm{s}^{\rm S2S}_{u};\bm{s}^{\rm LM}_{u}] + \bm{b}^{\rm g}) \\
& \bm{s}^{\rm CF}_{u} = \bm{W}^{\rm CF}[\bm{s}^{\rm S2S}_{u};\bm{g}_{u} \odot \bm{s}^{\rm LM}_{u}] + \bm{b}^{\rm CF} \\
& P_{\rm S2S}(\bm{y}|\bm{x}) = {\rm softmax}({\rm ReLU}(\bm{W}^{\rm out}\bm{s}^{\rm CF}_{u} + \bm{b}^{\rm o}))
\end{align*}
where $\bm{W}^{*}$ and $\bm{b}^{*}$ are trainable parameters, $\bm{d}_{u}^{\rm LM}$ is a hidden state of RNNLM, $\bm{s}_{u}^{\rm LM}$ is a feature from the external LM, $\bm{s}_{u}^{\rm CF}$ is a bottleneck feature before the final softmax layer, $\bm{g}_{u}$ is a gating function, and $\odot$ represents element-wise multiplication.
ReLU non-linear function is inserted before the softmax layer as suggested in \cite{cold_fusion}.
We use the hidden state as a feature from RNNLM instead of logits because we use the universal character vocabulary for multilingual experiments, which results in the large softmax layer and increases the computational time \cite{toshniwal2018comparison}.

In the original formulation in \cite{cold_fusion,toshniwal2018comparison}, scores from the external LM are not used.
We found that linear interpolation of log probabilities from the LM with those from the S2S model during the inference as in shallow fusion still has complementary effects to improve performance.
Therefore, we adopt it in all experiments.

\subsubsection{Deep fusion (fine-tuning fusion)}\label{sssec:deep_fusion}
\textit{Deep fusion} \cite{deep_fusion} is another method to integrate an external LM during training.
Unlike cold fusion, deep fusion is applied only for fine-tuning the gating part after parameters of both the pre-trained S2S model and RNNLM are frozen.
Although deep fusion is formulated with a scalar gating function in \cite{deep_fusion}, we use the same architecture as cold fusion in Section \ref{sssec:cold_fusion} to make a strict comparison.
Then, the difference from the cold fusion is in the timing to integrate the external LM (from scratch or in the middle stage) and which parameters to update after integration (see Figure \ref{fig:fusion_transfer}).

\section{Transfer learning of multilingual ASR}\label{sec:trainser_learning} 

\subsection{Adaptation to a target language}
We adapt a seed language-independent end-to-end ASR model to an (unseen) target language.
We investigate the following four scenarios: 
\begin{description}
\item {\bf \em multi10:} From non-target 10 languages to an unseen target language
\item {\bf \em high2:}  From 2 high resource languages (English and Japanese) to an unseen target  language 
\item {\bf \em multi10+high2:} From the mix of non-target 10 languages and 2 high resource languages to an unseen target language
\item {\bf \em multi15:} From the mix of non-target 10 languages and target 5 languages to a particular target language
\end{description}
The top three conditions are regarded as cross-lingual adaptation.

\subsection{LM fusion transfer}\label{ssec:lm_fusion_transfer}
During adaptation, all parameters are copied from the seed language-independent S2S model, then training is continued toward a target language.
We investigate improved adaptation methods by integrating the external LM during and/or after transfer learning from the seed model.
Three methods are considered as follows:
\begin{description}
\item {\bf Transfer + SF: } Shallow fusion in Section \ref{sssec:shallow_fusion} is conducted in the inference stage after adaptation.
\item {\bf Cold fusion transfer (CF-transfer): } Cold fusion in Section \ref{sssec:cold_fusion} is conducted during adaptation. We integrate the external RNNLM from the start point of adaptation to a target language. The softmax layer is randomly initialized before adaptation due to the additional gating part.
\item {\bf Deep fusion transfer (DF-transfer):} Deep fusion in Section \ref{sssec:deep_fusion} is conducted after adaptation. DF-transfer is composed of two stages: (1) adaptation by updating the whole parameters until convergence, and (2) fine-tuning only the gating part after integrating the external RNNLM. The softmax layer is randomly initialized before stage (2).
\end{description}


\section{Experimental evaluation}\label{sec:exp}
\subsection{Experimental setting}
We used data from the IARPA BABEL project \cite{babel} and selected 10 languages as non-target languages for training the seed language-independent model: Cantonese, Bengali, Pashto, Turkish, Vietnamese, Haitian, Tamil, Kurmanji, Tokpisin and Georgian, and 5 languages for adaptation: Assamese (AS), Swahili (SW), Lao (LA), Tagalog (TA) and Zulu (ZU).
Full language pack (FLP) is used for all experiments except for Section \ref{sssec:result_lm_fusion_transfer}, where limited language pack (LLP) which consists of about 10\% of FLP is used for adaptation.
We sampled 10\% of data  from the training data for each language as the validation set.
In addition, we used Librispeech corpus \cite{librispeech} and the Corpus of Spontaneous Japanese (CSJ) \cite{csj} as additional high resources.

We used Kaldi toolkit \cite{kaldi} for feature extraction.
The input features were static 80-channel log-mel filterbank outputs appended with 3-dimensional pitch features computed with a 25ms window and shifted every 10ms.
The features were normalized by the mean and the standard deviation on the whole training set.
For the vocabulary, we used the universal character set including all characters from all languages \cite{watanabe2017language}, resulting in the vocabulary size of 5,353 classes including 17 language IDs, \textit{sos}, \textit{eos}, \textit{unk}, and blank labels.
For multilingual experiments, we prepended the corresponding language ID so that the decoder network can jointly identify the correct target language while recognizing speech \cite{watanabe2017language}.

Our encoder network is composed of two VGG-like CNN blocks \cite{vgg} followed by a max-pooling layer with a stride of 2 $\times$ 2, and 5 layers of bidirectional long short-term memory (BLSTM) \cite{lstm} with 1024 memory cells, which results in time reduction by a factor of 4.
The decoder network consists of two layers of LSTM with 1024 memory cells.
For both monolingual and multilingual experiments, we used the same architecture.
Training was performed on the mini-batch size of 15 utterances using Adadelta \cite{adadelta} algorithm with an initial epsilon $1e-8$.
Epsilon was divided by a factor of 0.01 when the teacher-forcing accuracy does not improve for the validation set at each epoch.
Scheduled sampling \cite{scheduled_sampling} with probability 0.4 and dropout for the encoder network with probability 0.2 were performed in all experiments during adaptation.
We set the CTC weight during training and decoding to 0.5 and 0.3, respectively.
We also set the beam width to 20 and the LM weight to 0.3. 

For RNNLM, we used two layers of LSTM with 650 memory cells.
All RNNLMs were trained with transcriptions in the parallel data except for experiments in Table \ref{table:result_lm_fusion_transfer_llp}.
We used stochastic gradient descent (SGD) for RNNLM optimization.
All networks are implemented by ESPnet toolkit \cite{espnet} with pytorch backend \cite{pytorch}.

\subsection{Results}\label{ssec:results}

\begin{table}[!t]
  \centering
  \begingroup
  \caption{Results of baseline monolingual systems. None of adaptation methods is conducted.}
   \label{table:result_mono}
   \vspace{-3mm}
  \begin{tabular}{l|ccccc} \hline
  \multirow{3}{*}{\hspace{6mm}Model} & \multicolumn{5}{c}{WER (\%)} \\ \cline{2-6}
    & AS & SW & LA & TA & ZU  \\
        & (54h) & (39h) & (58h) & (75h) & (54h) \\ \hline
      Old baseline \cite{jj_slt18} & 73.9 & 66.5 & 64.5 & 73.6 & 76.4  \\ 
      New baseline & 64.5 & 56.6 & 56.2 & 56.4 & 69.5  \\ 
      \ + large units & 59.9 & 50.9 & 51.7 & 52.7 & 65.5  \\ 
      \ + shallow fusion & {\bf 57.4} & {\bf 46.5} & {\bf 49.8} & {\bf 49.9} & {\bf 62.9} \\ \hline
      BLSTM-HMM & 49.1 & 38.3 & 45.7 & 46.3 & 61.1  \\ \hline
    \end{tabular}
    \vspace{-4mm}
  \endgroup
\end{table}

\subsubsection{Baseline monolingual systems for target 5 languages}\label{sssec:result_mono}
First, we present the results of the baseline monolingual end-to-end systems in Table \ref{table:result_mono}.
Our new systems (line 2) significantly outperformed the old baseline reported on \cite{jj_slt18}.
The gain mostly came from adding VGG blocks before BLSTM encoder and one more decoder LSTM layer though we also tuned other hyper-parameters.
Next, changing the unit sizes of each LSTM layer from 320 to 1024 drastically improved the performance.
This is surprising because increasing the number of parameters often makes the model overfit to the small amount of training data.
Finally, shallow fusion with the monolingual RNNLM further boosted the performance although the RNNLM was trained with the small amount of transcriptions only.
We use this setting as default in the rest of experiments.

We also built BLSTM-HMM hybrid systems for comparison.
The BLSTM-HMM architecture includes 3 BLSTM layers each with 512 memory cells and 300 projection units\footnote{Increasing the unit size did not lead to any improvement.}.
The BLSTM acoustic model was trained using the latency control technique with 22 past frames and 21 future frames.
The acoustic model receives 40-dimensional filterbank features as input.
N-gram language model is built with the training transcriptions.
WERs by our end-to-end systems with shallow fusion are close to those of the hybrid system, just 3.6 and 1.8 \% absolute difference for Tagalog and Zulu, respectively.

\subsubsection{Comparison of seed language-independent models}\label{sssec:result_seed}
We compared the seed language-independent models for adaptation to target languages.
All models were transferred, and shallow fusion with the corresponding monolingual RNNLM trained with the parallel data was performed.
The results are shown in Table \ref{table:result_change_seed}.
The overall performance was significantly improved by transfer learning.
The transferred S2S models achieved comparable WER to BLSTM-HMM for Tagalog and outperformed for Zulu in Table \ref{table:result_mono}.
We can see that \textit{multi10} model is generally better than \textit{high2} model despite the smaller data size, and combination of them (\textit{multi10+high2}) gives slight improvement.
On the other hand, \textit{multi15} model that includes the target language does not lead to further improvement even after fine-tuning.
We can conclude that the diversity of languages is more important than the total amount of training data, and 10 languages are almost sufficient for learning language-independent feature representation and generalized to other languages well \cite{dalmia2018sequence}.
Since \textit{multi10} shows the competitive results to \textit{multi10+high2} only with one third training data, we use \textit{multi10} as the seed model and investigate cross-lingual adaptation in the following experiments.

\begin{table}[!t]
  \centering
  \begingroup
  \caption{Results of adaptation from the different seed language-independent models. Shallow fusion with the corresponding monolingual RNNLM was conducted.}
  \label{table:result_change_seed}
  \vspace{-3mm}
  \begin{tabular}{l|r|ccccc} \hline
  \multirow{2}{*}{\hspace{7mm}Seed} & \multirow{2}{*}{{\em hours}} & \multicolumn{5}{c}{WER (\%)} \\ \cline{3-7}
    &  & AS & SW & LA & TA & ZU  \\ \hline
	   multi10 & 643 & 53.4 & 41.3 & 46.1 & 46.4 & 60.2  \\
      high2 & 1,472 & 57.8 & 45.0 & 48.6 & 49.4 & 61.9  \\
      multi10+high2 & 2,115 & {\bf 53.2} & 40.7 & 45.1 & {\bf 45.3} & {\bf 58.5}  \\
      multi15 & 929 & 53.4 & {\bf 40.6} & {\bf 45.0} & 46.1 & 58.8 \\ \hline
      multi15 w/o FT & 929 & 56.2 & 44.2 & 47.1 & 47.8 & 60.6  \\ \hline
    \end{tabular}
    \vspace{1mm}
    \\ (FT: fine-tuning to a target language)
    \vspace{-4mm}
  \endgroup
\end{table}

\subsubsection{Effect of LM fusion transfer}\label{sssec:result_lm_fusion_transfer}
The results of our proposed LM fusion transfer are given in Table \ref{table:result_lm_fusion_transfer_flp}.
When training S2S models from scratch, there is no difference among all fusion methods.
When transferred from the language-independent S2S model, significant improvement is observed by integrating the external RNNLM.
Shallow fusion was more effective than when training the S2S models from scratch in Table \ref{table:result_mono} because the multilingual training led to generalization and the affinity for the external LM was enhanced.
CF-transfer got some improvements compared to transfer learning with shallow fusion for 3 target languages, but the effects of DF-transfer and CF-transfer are not significant.
This is because RNNLMs were trained with text in the small parallel data only, therefore linguistic context during adaptation was not so effective.
However, CF-transfer in Tagalog outperformed the monolingual hybrid system in Table \ref{table:result_mono}.
When compared to the previous work using the same data \cite{jj_slt18}, CF-transfer yielded 21.6\% gains relatively on average.
Furthermore, 6.8\% gains were achieved from transfer learning without the external LM.

\begin{table}[!t]
  \centering
  \begingroup
  \caption{Results of LM fusion transfer on FLP ($\sim$50h)}
  \label{table:result_lm_fusion_transfer_flp}
  \vspace{-3mm}
  \begin{tabular}{l|l|ccccc} \hline
  \multicolumn{2}{c|}{\multirow{2}{*}{Model}} & \multicolumn{5}{c}{WER (\%)} \\ \cline{3-7}
	\multicolumn{2}{c|}{} & AS & SW & LA & TA & ZU  \\ \hline
    \multirow{1}{*}{Transfer \cite{jj_slt18}} & SF & 65.3 & 56.2 & 57.9 & 64.3 & 71.1  \\  \hline
     \multirow{4}{*}{Scratch} & --- & 59.9 & 50.9 & 51.7 & 52.7 & 65.5  \\
       & SF & 57.4 & 46.5 & 49.8 & 49.9 & 62.9 \\ 
       & DF+SF & 57.5 & 46.4 & 49.9 & 49.9 & 62.6 \\ 
       & CF+SF & 57.5 & 47.3 & 50.0 & 50.2 & 62.9 \\ \hline
      \multirow{4}{*}{\shortstack{Transfer\\(multi10)}}& --- & 56.4 & 46.4 & 48.6 & 50.1 & 63.5 \\ 
      & SF & {\bf 53.4} & 41.3 & 46.1 & 46.4 & 60.2 \\
      & DF+SF & 53.5 & {\bf41.2} & 46.2 & {\bf 46.2} & 59.9 \\
      & CF+SF & 53.6 & 41.6 & {\bf 45.9} & {\bf 46.2} & {\bf 59.5} \\ \hline
    \end{tabular}
     \vspace{1mm}
    \\ (SF: shallow fusion, \ DF: deep fusion, \ CF: cold fusion)
    \vspace{2mm}
  \endgroup
\end{table}

To investigate the effect of additional text data, we evaluate the LM fusion transfer with LLP on each target language ($\sim$10 hours).
The results are shown in Table \ref{table:result_lm_fusion_transfer_llp}.
We used monolingual RNNLM trained with LLP (parallel data) and FLP ($\sim$50 hours), respectively.
The latter setting of a small speech data set ($\sim$10 hours) and a larger text data set ($\sim$50 hours) is regarded as a more realistic scenario in low-resource languages.
When training S2S models from scratch, all models could not converge in our implementation even when reducing the unit sizes.
The Babel corpus is mostly composed of conversational telephone speech (CTS), so it is difficult to optimize the S2S model from scratch with just around 10-hour training data.
In the transfer learning approach, all three fusion methods got significant gains by using the external LM except for deep fusion in Assamese.
For RNNLM trained with LLP, all fusion methods achieved a larger improvement than in Table \ref{table:result_lm_fusion_transfer_flp}.
Interestingly, WER significantly dropped even when each RNNLM was trained with 10-hour data only.
But all fusion methods show similar performance.
In contrast, CF-transfer significantly outperformed simple transfer learning with shallow fusion on all 5 target languages when the RNNLM was trained with FLP, which is five-times larger than LLP.
Therefore, we can conclude that linguistic context is helpful for adaptation when additional text data is available.
This shows CF-transfer in Table \ref{table:result_lm_fusion_transfer_flp} has the potential to surpass transfer learning with shallow fusion if we can access to additional text data\footnote{Since \textit{the provided data only} can be used for system training in BABEL rules, we do not explore to crawl text data from the WEB.}.
In summary, CF-transfer yielded relative 10.4\% and 2.3\% gains on average compared to transfer learning without and with shallow fusion, respectively.

\begin{table}[!t]
  \centering
  \footnotesize
  \begingroup
  \caption{Results of LM fusion transfer on LLP ($\sim$10h)}
  \label{table:result_lm_fusion_transfer_llp}
  \vspace{-3mm}
  \begin{tabular}{l|l|c|ccccc} \hline
  \multicolumn{2}{c|}{\multirow{3}{*}{Model}} & \multirow{3}{*}{\shortstack{LM\\data}} & \multicolumn{5}{c}{WER (\%)} \\ \cline{4-8}
	\multicolumn{2}{c|}{} &  & AS & SW & LA & TA & ZU  \\
   \multicolumn{2}{c|}{} &  & (8h) & (9h) & (9h) & (9h) & (9h) \\ \hline
     Scratch & --- & --- & \multicolumn{5}{c}{not converge}  \\ \hline
        \multirow{7}{*}{\shortstack{Transfer\\(multi10)}} & --- & --- & 67.5 & 59.7 & 60.3 & 66.2 & 75.4 \\ \cline{3-3} 
      & SF & \multirow{3}{*}{LLP}  & 63.3 & 52.8 & 57.2 & 60.8 & 71.2 \\
      & DF+SF &  & 68.0 & 52.4 & 57.3 & 60.7 & 70.9 \\
      & CF+SF &  & 63.2 & 52.8 & 58.4 & 60.6 & 71.0 \\ \cline{2-8}
      & SF & \multirow{3}{*}{FLP} & 62.7 & 51.7 & 56.4 & 60.0 & 71.0 \\
      & DF+SF &  & 66.8 & 50.7 & 56.1 & 60.0 & 69.9 \\
      & CF+SF &  & {\bf 61.7} & {\bf 50.3} & {\bf 56.0} & {\bf 57.9} & {\bf 69.8} \\ \hline 
    \end{tabular}
    \vspace{-4mm}
  \endgroup
\end{table}

\section{Conclusion}\label{sec:conclusion}
We explored the usage of linguistic context from the external LM during adaptation of the language-independent S2S model to target low-resource languages.
We empirically compared various LM fusion methods and confirmed their effectiveness in resource limited situations.
We showed that cold fusion transfer is more effective than simply applying shallow fusion after adaptation when additional text is available, which means linguistic context is also helpful in addition to acoustic adaptation.
Our S2S model drastically closed the gap from the BLSTM-HMM hybrid system.



\footnotesize
\bibliographystyle{IEEEbib}
\bibliography{refs}

\end{document}